\documentclass[runningheads]{llncs}

\usepackage{eccv}

\usepackage{eccvabbrv}
\usepackage{pdfpages}
\usepackage{graphicx}
\usepackage{booktabs}

\usepackage[accsupp]{axessibility} 

\usepackage{array}

\newcolumntype{P}[1]{>{\centering\arraybackslash}p{#1}}
\newcommand{\norm}[1]{\left\lVert#1\right\rVert}
\usepackage{orcidlink}

\begin{document}

% ---------------------------------------------------------------
\title{Open-Set Plankton Recognition} 

\author{Joona Kareinen \inst{1}\orcidlink{0009-0005-5997-2327}\and
Annaliina Skyttä \inst{2} \and
Tuomas Eerola \inst{1}\orcidlink{0000-0003-1352-0999} \and
Kaisa Kraft \inst{2}\orcidlink{0000-0001-6290-3887} \and
Lasse Lensu \inst{1}\orcidlink{0000-0002-7691-121X} \and
Sanna Suikkanen \inst{2}\orcidlink{0000-0002-0768-8149} \and
Maiju Lehtiniemi \inst{2}\orcidlink{0000-0003-4782-4958} \and
Heikki Kälviäinen \inst{1}\orcidlink{0000-0002-0790-6847}
}

\authorrunning{J.~Kareinen et al.}

\institute{Computer Vision and Pattern Recognition Laboratory,
Lappeenranta-Lahti University of Technology LUT, Lappeenranta, Finland \and
Finnish Environment Institute, Helsinki, Finland}

\maketitle
 
\begin{abstract}
    This paper considers open-set recognition (OSR) of plankton images. Plankton include a diverse range of microscopic aquatic organisms that have an important role in marine ecosystems as primary producers and as a base of food webs. Given their sensitivity to environmental changes, fluctuations in plankton populations offer valuable information about oceans' health and climate change motivating their monitoring. Modern automatic plankton imaging devices enable the collection of large-scale plankton image datasets, facilitating species-level analysis. Plankton species recognition can be seen as an image classification task and is typically solved using deep learning-based image recognition models. However, data collection in real aquatic environments results in imaging devices capturing a variety of non-plankton particles and plankton species not present in the training set. This creates a challenging fine-grained OSR problem, characterized by subtle differences between taxonomically close plankton species. We address this challenge by conducting extensive experiments on three OSR approaches using both phyto- and zooplankton images analyzing also on the effect of the rejection thresholds for OSR. The results demonstrate that high OSR accuracy can be obtained promoting the use of these methods in operational plankton research. We have made the data publicly available to the research community.
\end{abstract}

\section{Introduction} \label{sec:intro}
Most image recognition models are trained based on the closed-set assumption where the test set is expected to contain the same set of classes as the data the model was trained on. However, this assumption does not hold in most real-world applications where the recognition model can be expected to encounter images from various previously unseen categories. This leads to an open-set scenario where the recognition model should be able to not only recognize the known classes but also identify if an image is from a previously unseen class. A good example of this is plankton species recognition from images. Automatic plankton imaging instruments capture various non-plankton particles when deployed in real aquatic environments. Moreover, plankton species vary between geographic regions, leading to datasets with only partly uniform class composition. To advance the use of plankton recognition models operationally, the open-set recognition  (OSR) problem should be addressed.

Plankton are microscopic organisms that live in aquatic environments, drifting and swimming along tides and currents. Plankton consists of two main groups: phytoplankton and zooplankton~\cite{flynn2019}. Phytoplankton, which obtains energy through photosynthesis, produces oxygen and organic compounds, which serve as an essential food source for various microorganisms. On the other hand, zooplankton are a primary food source for many animals and have a key role in nutrient cycling by consuming phytoplankton and transferring energy to higher trophic levels, such as commercially important fish species~\cite{zoo2023}. Plankton is vital for the marine food web as primary producers and indicators of ocean health, responding quickly to changes in temperature and currents~\cite{fenchel1988marine, hays2005}. Monitoring how plankton populations change can provide valuable information about the effects of climate change~\cite{hays2005}, driving the increasing collection of information about plankton concentrations and species distribution.

Recent technological advancements have led to automatic and semi-automatic plankton imaging systems, used to gather massive plankton datasets~\cite{global2019}. Going manually through these images is infeasible, and to fully utilize the available data, an automatic system recognizing plankton species with minimal supervision is required. Several challenges exist in plankton recognition including subtle visual differences between species, varying image quality, unrecognizable particles, and class imbalance~\cite{eerola2023survey}.

A large number of plankton recognition methods are available~\cite{eerola2023survey}. These vary from traditional feature engineering-based approaches~\cite{Tang1998,Luo2003} to convolutional neural networks~\cite{lumini2019,burevs2021plankton,kraft2022towards}, and lately, vision transformers~\cite{kyathanahally2022ensembles}. Despite the high recognition accuracies reported in studies, these methods have not been widely adopted for operational use. One notable reason for this is the open-set nature of the recognition task, which is not properly considered in most studies. Works that do propose a method for open-set plankton recognition~\cite{badreldeen2022open,pu2021anomaly} report notable drops in recognition accuracy compared to closed-set recognition models. Plankton image data introduces various challenges, such as the fine-grained nature of the recognition task and label uncertainty. These factors render the open-set plankton recognition problem more difficult than many other domains in which OSR methods have been traditionally developed. However, this also makes plankton recognition an interesting testing ground for developing general-purpose OSR methods.

\begin{figure}[tb]
    \centering
    \includegraphics[width=0.9\linewidth]{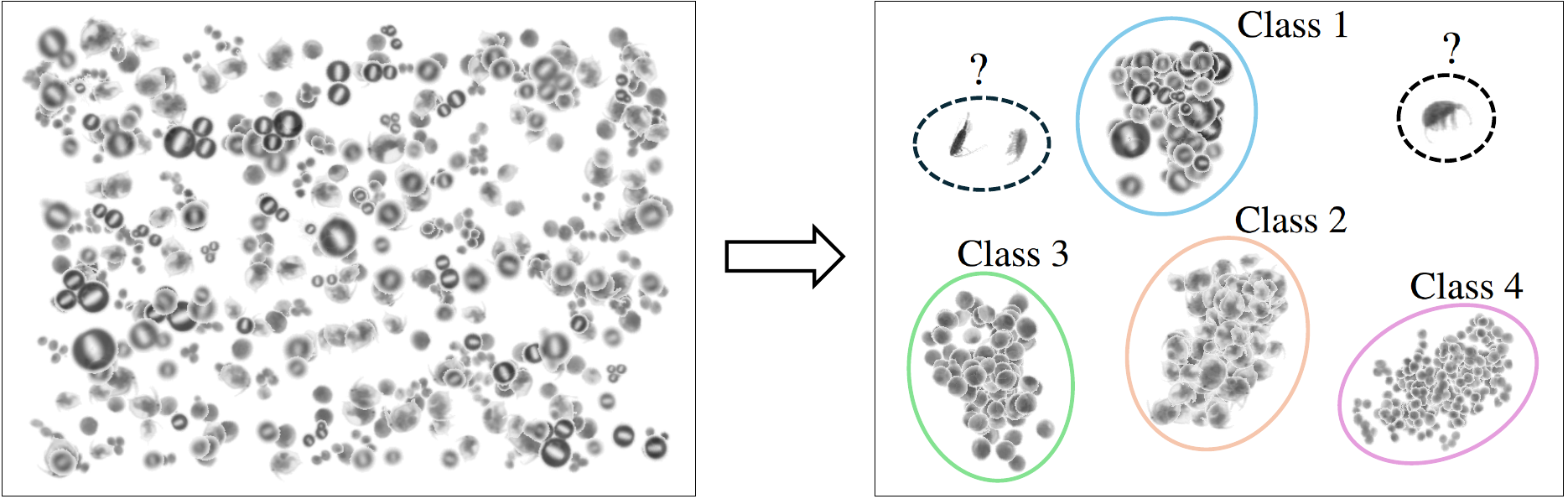}
    \caption{Open-set plankton recognition using metric learning. Classes 1–4 represent the known classes, and ? unknown classes.}
    \label{fig:plankton}
\end{figure}

In this paper, we conduct extensive experiments on three different OSR approaches for both phyto- and zooplankton images. The methods selected for the study are: (1) OpenMax~\cite{bendale2016towards} allowing the identification and handling of unknown classes by modifying the output probabilities, and two metric learning approaches: (2) ArcFace~\cite{deng2019arcface} and (3) Class Anchored Clustering (CAC)~\cite{miller2021class}. Metric learning (see Fig.~\ref{fig:plankton}) allows mapping images to a metric space where images from the same class are close to each other, enabling the recognition and identification of unknown classes simultaneously by setting distance (rejection) thresholds. For ArcFace, we utilize the implementation from~\cite{badreldeen2022open}, developed for open-set plankton recognition. CAC has not been previously applied to plankton recognition.

Due to the subtle differences between plankton species, the rejection thresholds have a crucial role in OSR as they form the basis for identifying unknown classes. 
We consider two approaches for selecting the rejection threshold values: (1) setting a fixed threshold for all classes based on validation set, and (2) class-specific thresholds.
To obtain class-specific threshold values, we utilize a quantile-based approach that does not require examples from unknown classes and can be used to find thresholds for all the classes based on a single quantile value minimizing the number of method hyperparameters. Furthermore, we study the sensitivity of the method to the threshold values. In the experimental part, we show that while the class-specific threshold values increase the accuracy of the methods on average, high accuracy can be obtained with a fixed threshold value when selected properly.

The main contributions of this study are as follows: (1) a new publicly available SYKE-plankton\_ ZooScan\_2024 zooplankton dataset~\cite{zooplanktondata} that together with the existing SYKE-plankton\_ IFCB\_2022 phytoplankton dataset \cite{kraft2022} provides a versatile and challenging dataset to develop and test plankton recognition and OSR methods, (2) the first application of CAC for open-set plankton recognition, and (3) extensive experiments on three OSR methods on plankton data including analysis of the effect of rejection thresholds.

\section{Related work}
\subsection{Plankton recognition}
A large number of automatic plankton recognition methods have been proposed to reduce manual work and to analyze the vast amount of plankton data. Early plankton recognition methods relied on classification techniques using handcrafted features. These feature engineering methods often combined various types of features (\eg, shape, texture, and local features) to obtain more representative feature vectors, and traditional classifiers such as support vector machines (SVMs) or decision trees to predict the class labels. For example, one of the first plankton recognition methods used moment invariants, Fourier descriptors, and granulometries as features, and employed a learning vector quantization artificial neural network as a classifier~\cite{Tang1998}. Luo \etal \cite{Luo2003} proposed an SVM-based classifier using global and domain-specific features.

During the recent decade, convolutional neural networks (CNNs) have become the dominant approach for plankton recognition. Various CNN architectures, including custom-designed ones, have been proposed, significantly improving the classification accuracy over traditional feature engineering methods. For example, Oord \etal \cite{deepsea2015} proposed a combination of a CNN inspired by VGGNet and a fully connected neural network. Orenstein \etal \cite{orenstein2017transfer} applied CNN-based features combined with a random forest classifier for phytoplankton images. Lumini and Nanni \cite{lumini2019} tested various pre-trained CNN architectures for plankton recognition, exploring both individual and fused networks. Kraft \etal \cite{kraft2022towards} studied operational phytoplankton recognition using ResNet-based plankton classifiers, and Batrakhanov \etal \cite{batrakhanov2024daplankton} explored the possibilities of adapting existing CNN-based models trained on one dataset to new datasets via domain adaptation methods. In recent years, Vision Transformers (ViTs) have increased in popularity, outperforming CNNs on various plankton datasets \cite{kyathanahally2022ensembles, maracani2023domain}. For a comprehensive survey on automatic plankton recognition, see~\cite{eerola2023survey}.

\subsection{Open-set recognition}
In closed-set recognition methods, an underlying assumption is that both training and testing data are drawn from the same label and feature space \cite{geng2021}. This means that the methods are only tested with known classes seen during training. Open-set recognition focuses on a more realistic scenario, where new classes, unseen during training, can appear in testing. In such a case, the classifier must identify the known objects correctly and simultaneously handle the unknown classes effectively.

Scheirer \etal \cite{scheirer2013} introduced key concepts for open-set recognition: open-space risk and openness. Open-space risk is defined as a fraction of open space over the measured space, indicating that as a sample moves farther from the training samples, there is an increased risk that the sample comes from an unknown class. The greater the open-space risk, the higher the probability that a sample is from an unknown class. Openness, on the other hand, measures how open a problem is, with 0 corresponding to a closed-set problem and $>$0 corresponding to an open problem. The degree of openness provides a quantitative measure of the extent to which a recognition system needs to generalize beyond the observed classes. Both definitions are important for mathematically formalizing the open-set recognition problem. 

Formally, an open-set recognition problem can be defined by combining the open space risk $R_\mathcal{O}$ and empirical risk $R_\epsilon$ that is minimized when training a classifier. Open-set recognition is to find a measurable recognition function $f \in \mathcal{H}$, where  $f(x) > 0$ implies positive recognition, and $f$ is defined by minimizing the open set risk
\begin{equation}
    \arg \min\limits_{f \in \mathcal{H}} \{R_\mathcal{O}(f) + \lambda_r R_\epsilon(f(V))\},
\end{equation}
where $V$ is training data and $\lambda_r$ is a regularization term that controls the balance between these risks \cite{scheirer2013}.

Recent work by Vaze \etal \cite{vaze2022openset} highlighted a significant correlation between closed-set and open-set performance in models. Their results showed that even a simple method using a maximum logit score for open-set scoring can achieve results as good as, or even better than, more advanced OSR methods. This finding suggests that enhancing a model's closed-set accuracy can be a straightforward and effective strategy for improving open-set recognition, potentially simplifying the design of OSR methods.

In the plankton recognition context, the open-set problem is often formulated as anomaly detection, with the goal being to correctly classify known classes and to reject unknown samples \cite{eerola2023survey}. In anomaly detection, the sample is considered an anomaly in case its features significantly differ from the feature average of the class. Pu \etal \cite{pu2021anomaly} proposed a loss function for anomaly detection that combines the cross-entropy loss used in classification, Kullback-Leibler divergence, and the Anchor loss. The method was tested with a plankton dataset that included images with bubbles and other invalid particles.  

Pastore \etal \cite{pastore2022anomaly} proposed a novel anomaly detection method for plankton recognition named as the TailDeTect (TDT) algorithm. The TDT algorithm is trained to detect deviations from an average sample in the feature space. When such a deviation is detected, the sample is labeled as an anomaly. The approach involves training one TDT detector for each class in the training set. A sample is considered a global anomaly if all detectors classify it as an anomaly, with global anomalies stored for further investigation.

Yang \etal \cite{yang2022contrastive} introduced in situ plankton image retrieval (IsPlanktonIR) in which a feature extractor is trained using the Supervised Contrastive (SupCon) loss \cite{khosla2020supervised}. During testing, a gallery set of known classes is used, and the similarity metric between input images and the gallery set is computed. If the similarity exceeds a predefined threshold, the class is classified with the same label and otherwise rejected as unknown. Similarly, Badreldeen \etal \cite{badreldeen2022open} proposed a method utilizing a CNN trained using the Angular Margin loss (ArcFace) \cite{deng2019arcface}, also employing cosine similarity and a threshold to determine between known and unknown classes.

\section{Methods}
Three OSR methods were selected for the study. The first method utilizes the OpenMax layer~\cite{bendale2016towards} developed for the OSR. The second method is an open-set plankton recognition method~\cite{badreldeen2022open} that uses similarity learning and the ArcFace loss~\cite{deng2019arcface}. The third method is the Class Anchored Clustering based OSR method~\cite{miller2021class} that has not been previously applied to plankton recognition.

\subsection{OpenMax}
OpenMax, proposed by Bendale and Boult \cite{bendale2016towards}, addresses the challenge of deep neural networks (DNNs) making overconfident predictions when faced with unknown classes. Unlike the traditional softmax layer that assigns all input data to known classes, OpenMax introduces a mechanism to identify and handle unknown classes by modifying the output probabilities. 

OpenMax uses a base neural network $f$ that takes in an image input $\boldsymbol{x}$ to produce activation vector $\boldsymbol{z} = f(\boldsymbol{x})$. Using these activation vectors, a Mean Activation Vector (MAV) is calculated for each class by averaging the activation vectors of correctly classified samples from the training set, that is,
\begin{equation}
    \mu_j = \frac{1}{N_j}\sum_{i=1}^{N_j} z_{ij},
\end{equation}
where $N_j$ is the number of correctly classified samples for class $j$, and $z_{ij}$ are the corresponding activation vectors. After calculating the MAVs, the distances between each class MAV and the activation vectors of correctly classified samples are calculated, and a Weibull distribution is fitted to $\eta$ largest distances. The fitting process yields a model $\rho_j$ for each class, containing three parameters: $\tau_j$ for shifting the data, and the Weibull shape and scale parameters $\kappa_j$ and $\lambda_j$. 

\subsubsection{Classifying samples.}
When the MAVs and the Weibull parameters have been obtained, OpenMax can be used to classify samples. During inference, the activation scores are sorted in descending order $s(i) = \text{argsort}(z_j(x))$, and the $\alpha$ top class scores are adjusted. For these top classes, the scaling factors $\omega_j$ are computed, while $\omega_j$ is set to 1 for other classes. The adjustment factor $\omega$ for the $s(i)$-th highest activation vector is calculated using the Weibull cumulative distribution function as
\begin{equation}
    \omega_{s(i)}(x) = 1 - \frac{\alpha - i}{\alpha}\text{exp}{\left( \frac{\norm{x - \tau_{s(i}})}{\lambda_{s(i)}} \right)}^{\kappa_{s(i)}},
\end{equation}
where $s(i)$ is the index of the $i$-th highest activation score, with $i$ ranging from 1 to $\alpha$. Following this, the adjusted activation vectors are calculated as
\begin{equation}
    \hat{\boldsymbol{z}}(x) = \boldsymbol{z}(x) \circ \boldsymbol{\omega}(x),
\end{equation}
where $\circ$ denotes an element-wise product. Additionally, an activation vector for an unknown class is computed which quantifies the overall likelihood that the sample does not belong to any of the known classes as 
\begin{equation}
    \hat{z_0}(x) = \sum_i z_i(x)(1 - \omega_i(x)).
\end{equation}

The final classification decision is made by applying the softmax function to the adjusted activation vectors obtaining $\hat{P}(y = j |x)$ for each class $j$. The predicted class is the one with the highest probability $y^* = \text{argmax } (\hat{P}(y = j |x))$. However, if the predicted class corresponds to an unknown class (denoted as class $0$), or if the highest probability is less than a predefined threshold $\delta$, the sample is rejected as an unknown. 

\subsection{ArcFace}
Additive Angular Margin Loss (ArcFace) was originally proposed by Deng \etal \cite{deng2019arcface} to enhance the discriminative power of face recognition systems. We utilize the ArcFace-based OSR method for plankton recognition proposed by Badreldeen \etal \cite{badreldeen2022open}. ArcFace integrates an additive angular margin penalty into the softmax loss to improve the feature separability across classes.

\subsubsection{Additive Angular Margin loss.}
ArcFace uses a base neural network $f$ that takes an image input $\boldsymbol{x}$ and transforms it into a vector of class features (embeddings) $\boldsymbol{z} = f(\boldsymbol{x})$. It trains the network using the Additive Angular Margin loss, defined as
\begin{equation}
    \mathcal{L} = - \log \left(  \frac{e^{s(\cos(\theta_{y_i} + m))}}{ e^{s(\cos(\theta_{y_i} + m))} + \sum_{j=1, j \neq y_i}^N e^{s(\cos(\theta_j)} }  \right),
\end{equation}
where $s$ is the hypersphere radius, $\theta_{y_i}$ is the angle between the feature and the class center of the correct class $y_i$, $\theta_j$ is the angle between the weight vector for class $j$ and the predicted feature vector, $N$ is the number of classes, and $m$ is the additive margin penalty. The additive margin penalty is used to increase inter-class separability and decrease intra-class variation.

\subsubsection{Classifying samples.}
After training a CNN model using the Additive Angular Margin loss, the model outputs embedding vectors for images. The similarity between two images is quantified by using the cosine distance
\begin{equation}
    d_{\cos}(\boldsymbol{z_1}, \boldsymbol{z_2}) = \frac{\boldsymbol{z_1} \cdot \boldsymbol{z_2}}{\norm{\boldsymbol{z_1}}\norm{\boldsymbol{z_2}}},
\end{equation}
where $\boldsymbol{z_1}$ and $\boldsymbol{z_2}$ are the embedding vectors. 

To classify images, a balanced gallery set is created by selecting example images for each class and calculating their embeddings. For a query image, the embedding is obtained by passing the image through the CNN. The cosine similarities between this embedding and those in the gallery set are then computed, and the query image is labeled based on the most similar image. If none of the similarities is higher than a predefined threshold $\delta$, the sample is rejected as unknown. 

\subsection{Class Anchored Clustering}
The Class Anchored Clustering (CAC), proposed by Miller \etal~\cite{miller2021class}, involves training a CNN with the CAC loss to cluster known classes around anchored class-dependent centers in the feature space, enhancing robustness against inputs from unknown classes. 

CAC uses a base neural network $f$, that takes an image input $\boldsymbol{x}$ and transforms it into a vector of class features $\boldsymbol{z} = f(\boldsymbol{x})$. The method also involves a set of pre-defined non-trainable class centers $(\boldsymbol{c}_1, \boldsymbol{c}_2, \ldots, \boldsymbol{c}_N) \in \boldsymbol{C}$, with each center corresponding to one of the $N$ classes. Distances between the feature vectors $\boldsymbol{z}$ and each of the class centers can be calculated as 
\begin{equation}
    \boldsymbol{d} = (\norm{\boldsymbol{z} - \boldsymbol{c}_1}_2, \norm{\boldsymbol{z} - \boldsymbol{c}_2}_2, \ldots, \norm{\boldsymbol{z} - \boldsymbol{c}_N}_2)^\intercal.
\end{equation}

In CAC, the class centers are anchored in the feature space without the need to learn them.  For each of the $N$ classes, the class center $\boldsymbol{c}_i$ is anchored along its respective class coordinate axis in the $N$-dimensional feature space, formulated as 
\begin{align}
    \boldsymbol{C} = (\boldsymbol{c}_1, \boldsymbol{c}_2, \ldots, \boldsymbol{c}_N) = \alpha \cdot (\boldsymbol{e}_1, \boldsymbol{e}_2, \ldots, \boldsymbol{e}_N) \\
    \boldsymbol{e}_1 = (1, 0, \ldots, 0)^\intercal, \ldots, \boldsymbol{e}_N = (0, 0, \ldots, 1)^\intercal,
\end{align}
where $\alpha$ is the magnitude of the anchor center hyperparameter. After the model has been trained, the class centers are updated to match the mean position of the correctly classified samples.

\subsubsection{Class Anchored Clustering loss.}
The CAC loss \cite{miller2021class} serves two primary purposes: a) minimizing the distance between training inputs and their respective class center, and b) maximizing the distance to all the other class centers. CAC loss is defined as a combination of a modified Tuplet Loss $\mathcal{L}_\text{T}$ and an Anchor loss $\mathcal{L}_\text{A}$, weighted by a hyperparameter $\lambda$ as  
\begin{equation}
    \mathcal{L}_{\text{CAC}}(\boldsymbol{x}, y) = \mathcal{L}_\text{T}(\boldsymbol{x},y) + \lambda \mathcal{L}_\text{A}(\boldsymbol{x},y).
\end{equation}
 The modified Tuplet Loss is based on the works of Sohn \cite{sohn2016improved}, and it encourages each input sample $\boldsymbol{x}$ to maximize the distance between its correct class center and all other class centers defined as 
 \begin{equation}
    \mathcal{L}_\text{T}(\boldsymbol{x},y) = \log \left(\sum_{j = 1}^N e^{d_y - d_j} \right),
\end{equation}
where $d_y$ is the distance between the input $\boldsymbol{x}$ and its correct class center, and $d_j$ goes over the distance between $\boldsymbol{x}$ and all class centers. The Anchor loss term $\mathcal{L}_\text{A}$ ensures that each sample minimizes the distance to its correct class center and is defined as
\begin{equation}
    \mathcal{L}_\text{A}(\boldsymbol{x},y) = d_y = \norm{f(\boldsymbol{x}) - \boldsymbol{c}_y}_2.
\end{equation}

\subsubsection{Classifying samples.}
To perform plankton recognition for an input image, the input is first passed through the trained CNN to obtain the feature vector. Next, the distance between the feature vector and each class center is calculated. The rejection score is then calculated as the element-wise product of the distance vector and its inverted softmin as
\begin{equation}
    \boldsymbol{\gamma}= \boldsymbol{d} \circ (1 - \text{softmin}(\boldsymbol{d})).
\end{equation}
This follows as known classes have two properties: a) a high softmin score corresponding to the correct class center, and b) a low absolute distance to that center. The rejection score can be seen as the classifiers' disbelief that the input belongs to each known class. 

Once the rejection scores have been calculated for all classes, the input image is assigned to the class with the lowest rejection score. If the lowest rejection score is higher than a predefined threshold $\delta$, the sample is rejected as unknown.

\subsection{Rejection threshold}
\label{sec:threshold}
% From CAC
The rejection threshold determines when a query image is too dissimilar to known classes and is recognized as unknown. In metric learning-based OSR methods, it is a crucial parameter for balancing both known and unknown class accuracy. Typically, a single threshold is used for all classes due to its ease of optimization. However, class-specific thresholds can be more suitable due to potential cluster variations in the feature space between classes \cite{geng2021}.

We propose to use a quantile-based approach to define class-specific thresholds.
First, the rejection or similarity scores are calculated for each class. These depend on the type of the used model with probabilities for OpenMax, cosine similarities for ArcFace, and rejection scores for CAC. The threshold for each class $\delta_j$ is then set by taking a quantile of the scores for each class. The quantile $q_\alpha$ represents the score below which $\alpha \times 100\%$ of the class's scores fall.
With methods where a small score is better, the quantile should be set relatively large so that most known samples' scores fall below it, whereas if a large score is better, a low quantile is preferred.

The benefits of this approach are simplicity compared to methods like grid search and the possibility to define the threshold values based on known samples only. Moreover, since thresholds for all classes are defined based on a single quantile value the number of tunable parameters remains low compared to separately tuning each threshold value.

\section{Experiments}
\subsection{Data}
We utilize two different plankton datasets captured with different imaging instruments and consisting of different types of plankton (zooplankton and phytoplankton). The first dataset is the SYKE-plankton\_ZooScan\_2024 dataset~\cite{zooplanktondata} consisting of 24\,123 single-specimen images of zooplankton, divided into 20 classes. The dataset was acquired using the ZooScan instrument~\cite{zoo2023} and water samples collected from the Baltic Sea.  

Due to the varying rarity of the plankton species, the dataset is highly imbalanced, with the number of images per class ranging from 5 to 9\,322. Besides zooplankton, the data also contains non-plankton particles including bubbles and fish eggs. 

\begin{figure}[t]
    \centering
    \includegraphics[width=0.9\linewidth]{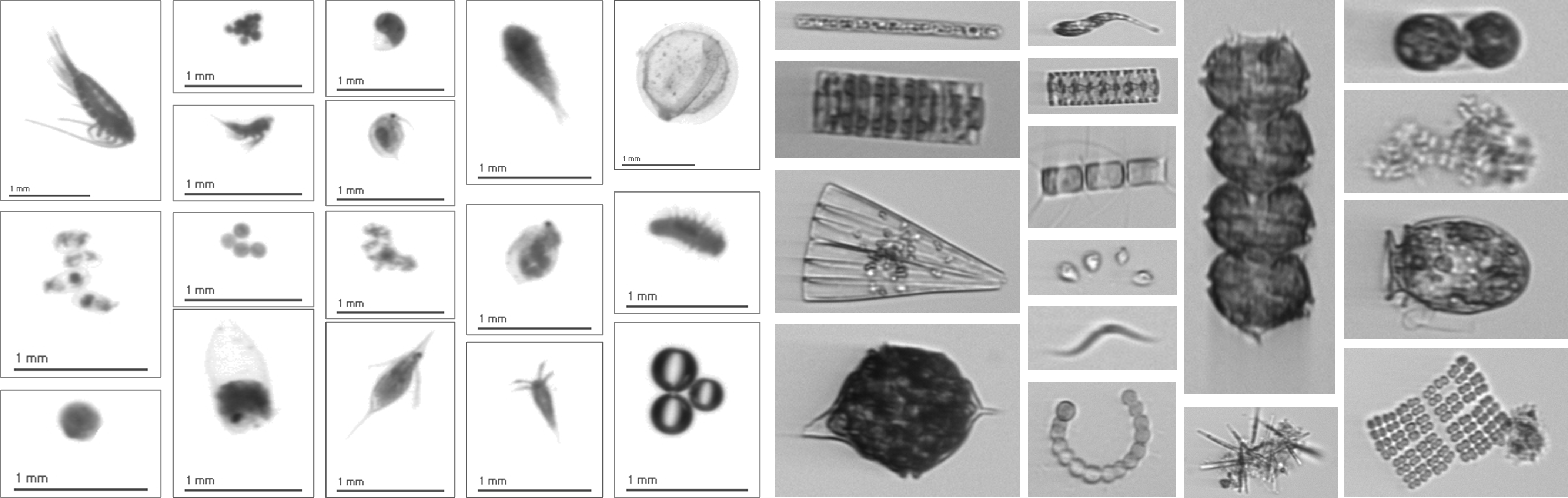}\\
    (a)\hspace{5cm}(b)
    \caption{Example plankton images: (a) SYKE-plankton\_ZooScan\_2024; (b) SYKE-plankton\_IFCB\_2022.}
    \label{fig:planktonImages}
\end{figure}

The second dataset is the SYKE-plankton\_IFCB\_2022 dataset \cite{kraft2022}, consisting of 63\,074 single-specimen images of phytoplankton captured in the Baltic Sea with the IFCB instrument \cite{flowcytobot}. These images are categorized into 50 different classes, with the number of images per class varying from 19 to 12\,280. Example images from both datasets can be seen in Fig.~\ref{fig:planktonImages}.

Datasets were split class-wise into training, validation, and testing sets in a 6:2:2 ratio. To reduce the effect of the class imbalance, training data were augmented so that each class contained 1000 samples. For classes with over 1000 samples, a random subset was used. For classes with fewer samples, the images were augmented with slight distortion and affine transformations (rotation, translation, rescaling, shearing). Additionally, all images were resized to a standard dimension of 224×224 pixels, preserving the aspect ratio.

\subsection{Description of Experiments}

The experiments evaluated the OSR performance of the three methods on both datasets. For both datasets, contain classes were designated as unknown and excluded from training, while the remaining classes were used for training. For the zooplankton dataset, 3 classes were excluded as unknowns, and 12 classes were used for training. This setup was repeated 5 times, with each class designated as unknown once. In this dataset, the 5 smallest classes (fewer than 50 images each) were excluded from training and used as extra unknowns for the testing set due to their limited size. For the phytoplankton dataset, 10 classes were selected as unknowns, with 40 classes used for training, following the same 5-fold testing setup. 

The OpenMax method was applied with two CNN backbones, ResNet18 and DenseNet121, and the method parameters were set as follows: $\eta = 20$ and $\alpha = 5$. Training was done over 100 epochs using the stochastic gradient descent (SGD) optimizer, with a learning rate of 0.01 and momentum set to 0.9. 
ArcFace was tested with the ResNet18 backbone and trained for 100 epochs using the AdamW optimizer, with a learning rate of 5e-5. The scale and margin parameters for ArcFace were set to 2.39 and 0.95, respectively. The gallery set size was 100 images per class. The CAC method was tested with ResNet18 and DenseNet121 backbones, and both were being trained for 100 epochs using the SGD optimizer, with a learning rate of 0.01 and momentum of 0.9, with the anchor magnitude set to 10, and the anchor loss weight to 0.1. For all methods, the ResNet18 and DenseNet121 backbones were pre-trained with the ImageNet dataset \cite{russakovsky2015imagenet}.

In addition to the direct comparison of the three methods, experiments on setting the rejection thresholds were carried out. All the OSR methods were tested using both a fixed threshold for all classes and class-specific thresholds. The fixed threshold for each method and dataset was determined by maximizing the open-set F-score \cite{mendes2017nearest} on the validation set including known and unknown classes. 
The open-set F-score is a metric that is calculated only for known classes but takes into account false positives and false negatives from the unknowns for each class. Class-specific thresholds were selected as described in Sec.~\ref{sec:threshold}. The threshold and quantile that achieved the highest average open-set F-score on the validation set across the five experiments were then used with the test set.

\subsection{Results}

The results for the zooplankton dataset when using the same threshold for all classes are presented in Table~\ref{tab:metrics_zoo_one_th}, while Table~\ref{tab:metrics_zoo} shows the results with class-specific thresholds. The “Known Class Accuracy” column represents the traditional closed-set scenario where the training and testing sets include the same set of classes. All methods demonstrated very high accuracy, exceeding 95\%. The “Known Class Accuracy (Th)” shows results when a threshold is applied to filter out unknowns, but the testing set still includes only known classes. Although there is a slight drop in the accuracies due to the falsely recognized unknowns, it remains high for all methods. The “Unknown Class Accuracy (Th)” column indicates the percentage of previously unseen images correctly rejected as unknowns, \ie, binary classification accuracy between knowns and unknowns. The “Open-set Accuracy” column shows the overall weighted accuracy across known and unknown classes, and the last column shows the “Open-set F-score”. The results show that CAC obtains the highest open-set accuracy and open-set F-score. When the results between a single threshold and class-specific thresholds are compared, the class-specific thresholds do slightly improve the open-set accuracy and open-set F-score on average. Additionally, on average the class-specific thresholds reduce the standard deviation of the open-set accuracy and open-set F-score. 

\begin{table}[tb]
    \centering
    \caption{Mean performance metrics and standard deviations of baseline models on the SYKE-plankton\_ZooScan\_2024 test set, averaged over 5 experiments using the same threshold for all classes.}
    \label{tab:metrics_zoo_one_th}
    \resizebox{\textwidth}{!}{
    \begin{tabular}{|l|P{2.5cm}|P{2.5cm}|P{2.5cm}|P{2.5cm}|P{2.5cm}|}
        \hline
        Model &  Known Class Accuracy & Known Class Accuracy (Th) & Unknown Class Accuracy (Th) & Open-Set Accuracy & Open-Set F-score \\ \hline
        OpenMax (ResNet)    & 97.24 ± 0.23\% & 94.61 ± 0.59\% & 64.82 ± 18.62\% & 90.31 ± 2.42\% & 93.32 ± 2.81\% \\ \hline
        OpenMax (DenseNet)  & 96.72 ± 1.41\% & 96.03 ± 1.78\% & 63.68 ± 17.52\% & 90.80 ± 3.77\% & 93.66 ± 3.52\% \\ \hline
        ArcFace (ResNet)    & 98.70 ± 0.54\% & 93.80 ± 2.34\% & 72.38 ± 19.41\% & 91.39 ± 1.24\% & 94.20 ± 1.19\% \\ \hline
        CAC (ResNet)        & 99.41 ± 0.22\% & 96.10 ± 1.30\% & \textbf{75.91} ± 18.32\% & \textbf{93.73} ± 0.78\% & \textbf{95.84} ± 0.89\% \\ \hline
        CAC (DenseNet)      & \textbf{99.42} ± 0.19\% & \textbf{96.87} ± 1.19\% & 69.61 ± 17.28\% & 92.31 ± 2.88\% & 94.83 ± 2.42\% \\ \hline
    \end{tabular}}
\end{table}

\begin{table}[tb]
    \centering
    \caption{Mean performance metrics and standard deviations of baseline models on the SYKE-plankton\_ZooScan\_2024 test set, averaged over 5 experiments using the class-specific thresholds.}
    \label{tab:metrics_zoo}
    \resizebox{\textwidth}{!}{
    \begin{tabular}{|l|P{2.5cm}|P{2.5cm}|P{2.5cm}|P{2.5cm}|P{2.5cm}|}
        \hline
        Model &  Known Class Accuracy & Known Class Accuracy (Th) & Unknown Class Accuracy (Th) & Open-Set Accuracy & Open-Set F-score \\ \hline
        OpenMax (ResNet)    & 97.24 ± 0.23\% & 93.80 ± 0.60\% & 71.15 ± 17.30\% & 90.83 ± 2.12\% & 93.68 ± 2.36\% \\ \hline
        OpenMax (DenseNet)  & 96.72 ± 1.41\% & 95.12 ± 0.81\% & 69.43 ± 19.30\% & 92.14 ± 1.71\% & 94.71 ± 1.63\% \\ \hline
        ArcFace (ResNet)    & 98.70 ± 0.54\% & \textbf{96.20} ± 0.62\% & 64.37 ± 20.35\% & 91.89 ± 1.64\% & 94.40 ± 2.01\% \\ \hline
        CAC (ResNet)        & 99.41 ± 0.22\% & 96.04 ± 0.30\% & \textbf{73.50} ± 19.31\% & \textbf{93.66} ± 0.83\% & \textbf{95.82} ± 0.59\% \\ \hline
        CAC (DenseNet)      & \textbf{99.42} ± 0.19\% & 95.80 ± 0.28\% & 71.99 ± 19.74\% & 93.15 ± 0.87\% & 95.47 ± 0.82\% \\ \hline
    \end{tabular}}
\end{table}

\begin{figure}[tb]
    \centering
    \includegraphics[width=0.9\linewidth]{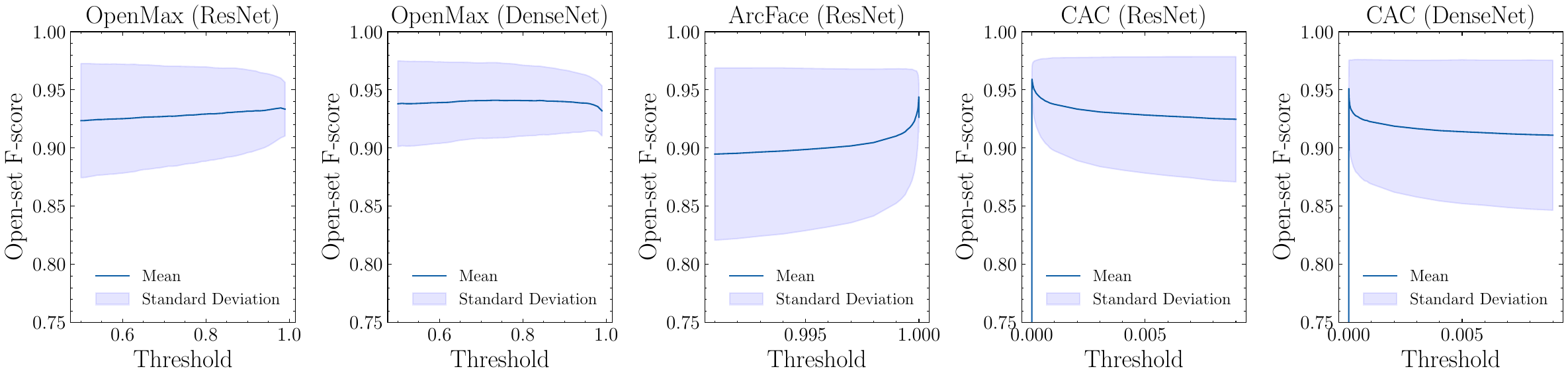}
    \caption{Average Open-set F-scores with standard deviations from 5 experiments on the SYKE-plankton\_ZooScan\_2024 validation set. The plots show the average performance of the baseline models across various thresholds.}
    \label{fig:thresholds_zoo_one_th}
\end{figure}

\begin{figure}[tb]
    \centering
    \includegraphics[width=0.9\linewidth]{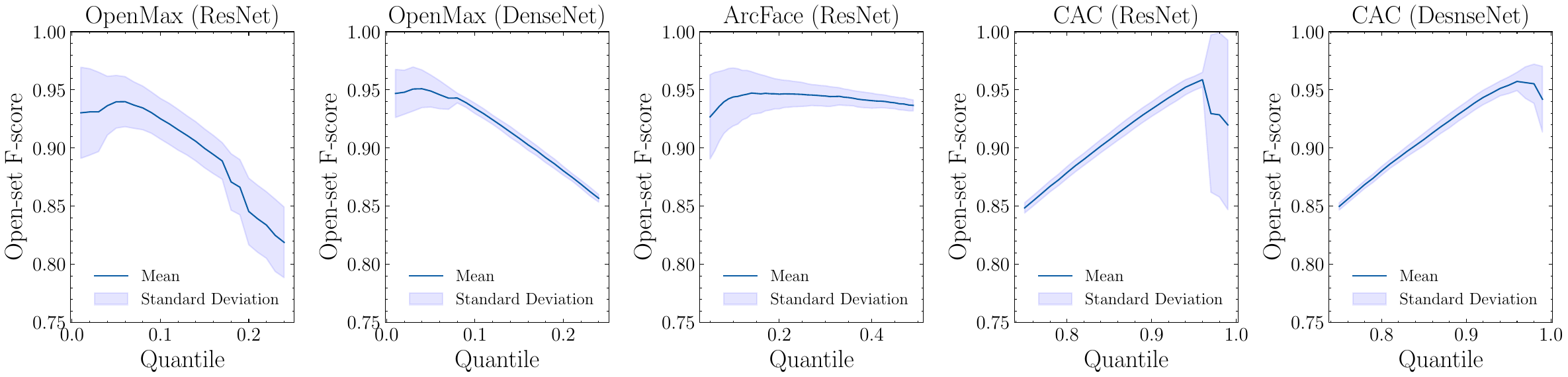}
    \caption{Average Open-set F-scores with standard deviations from 5 experiments on the SYKE-plankton\_ZooScan\_2024 validation set. The plots show the average performance of the baseline models across various quantiles.}
    \label{fig:thresholds_zoo}
\end{figure}

Figs.~\ref{fig:thresholds_zoo_one_th} and \ref{fig:thresholds_zoo} show how the threshold and quantile affect the open-set F-measure on the validation set. These figures show the mean and standard deviation over the 5 experiments. Fig.~\ref{fig:thresholds_zoo_one_th} shows the raw thresholds employed when applying a single threshold for all classes. For the OpenMax methods, there is no significant peak that optimizes the Open-set F-score. In contrast, the metric learning methods ArcFace and CAC show clear peaks, with optimal thresholds near 1 and 0, respectively. The threshold selection for CAC is less sensitive than for ArcFace as the Open-set F-score remains higher across the whole threshold range. With class-specific thresholds, there are clear optimal values for CAC and OpenMax. For ArcFace, the Open-set F-score is not very sensitive to the quantile selection.

The results for the phytoplankton dataset when using the same threshold for all classes are presented in Table~\ref{tab:metrics_phyto_one_th}, while Table~\ref{tab:metrics_phyto} shows the results with class-specific thresholds. When using the single threshold, CAC achieves the highest accuracies, except for a slightly higher unknown class accuracy obtained by ArcFace. In contrast, with the class-specific thresholds, ArcFace excels in the unknown class accuracy, open-set accuracy, and open-set F-score. Although the difference in open-set F-score between the CAC and ArcFace is not substantial, ArcFace achieves a significantly higher unknown class accuracy. Comparing the results between the single threshold and class-specific thresholds, the optimized single threshold yields better results on average.

\begin{table}[tb]
    \centering
    \caption{Mean performance metrics and standard deviations of baseline models on the SYKE-plankton\_IFCB\_2022 test set, averaged over 5 experiments using the same threshold for all classes.}
    \label{tab:metrics_phyto_one_th}
    \resizebox{\textwidth}{!}{
    \begin{tabular}{|l|P{2.5cm}|P{2.5cm}|P{2.5cm}|P{2.5cm}|P{2.5cm}|}
        \hline
        Model &  Known Class Accuracy & Known Class Accuracy (Th) & Unknown Class Accuracy (Th) & Open-Set Accuracy & Open-Set F-score \\ \hline
        OpenMax (ResNet)    & 97.01 ± 0.53\% & 93.06 ± 1.00\% & 71.63 ± 7.24\% & 88.95 ± 2.05\% & 92.61 ± 2.15\% \\ \hline
        OpenMax (DenseNet)  & 97.39 ± 0.77\% & 93.64 ± 1.33\% & 69.64 ± 12.80\% & 89.00 ± 3.31\% & 92.66 ± 2.79\% \\ \hline
        ArcFace (ResNet)    & 98.17 ± 0.56\% & 93.19 ± 1.80\% & \textbf{73.66} ± 10.16\% & 90.24 ± 1.98\% & 93.37 ± 1.61\% \\ \hline
        CAC (ResNet)        & \textbf{98.64} ± 0.51\% & \textbf{94.99} ± 1.00\% & 72.89 ± 8.12\% & \textbf{90.76} ± 1.61\% & \textbf{93.94} ± 1.59\% \\ \hline
        CAC (DenseNet)      & 98.52 ± 0.71\% & 94.41 ± 2.16\% & 68.61 ± 10.44\% & 89.44 ± 2.66\% & 93.11 ± 2.05\% \\ \hline
    \end{tabular}}
\end{table}

\begin{table}[tb]
    \centering
    \caption{Mean performance metrics and standard deviations of baseline models on the SYKE-plankton\_IFCB\_2022 test set, averaged over 5 experiments using class-specific thresholds.}
    \label{tab:metrics_phyto}
    \resizebox{\textwidth}{!}{
    \begin{tabular}{|l|P{2.5cm}|P{2.5cm}|P{2.5cm}|P{2.5cm}|P{2.5cm}|}
        \hline
        Model &  Known Class Accuracy & Known Class Accuracy (Th) & Unknown Class Accuracy (Th) & Open-Set Accuracy & Open-Set F-score \\ \hline
        OpenMax (ResNet)    & 97.01 ± 0.53\% & 93.27 ± 0.19\% & 67.10 ± 4.94\% & 88.01 ± 2.84\% & 91.94 ± 2.76\% \\ \hline
        OpenMax (DenseNet)  & 97.39 ± 0.77\% & \textbf{94.97} ± 0.45\% & 59.43 ± 12.19\% & 88.57 ± 2.45\% & 92.44 ± 2.30\% \\ \hline
        ArcFace (ResNet)    & 98.17 ± 0.56\% & 92.15 ± 0.83\% & \textbf{80.07} ± 7.39\% & \textbf{90.06} ± 1.19\% & \textbf{93.34} ± 1.11\% \\ \hline
        CAC (ResNet)        & \textbf{98.64} ± 0.51\% & 93.33 ± 0.32\% & 70.17 ± 7.68\% & 89.40 ± 0.99\% & 93.07 ± 1.31\% \\ \hline
        CAC (DenseNet)      & 98.52 ± 0.71\% & 93.18 ± 0.19\% & 73.55 ± 7.12\% & 89.65 ± 1.35\% & 93.15 ± 1.71\% \\ \hline
    \end{tabular}}
\end{table}

\begin{figure}[tb]
    \centering
    \includegraphics[width=0.9\linewidth]{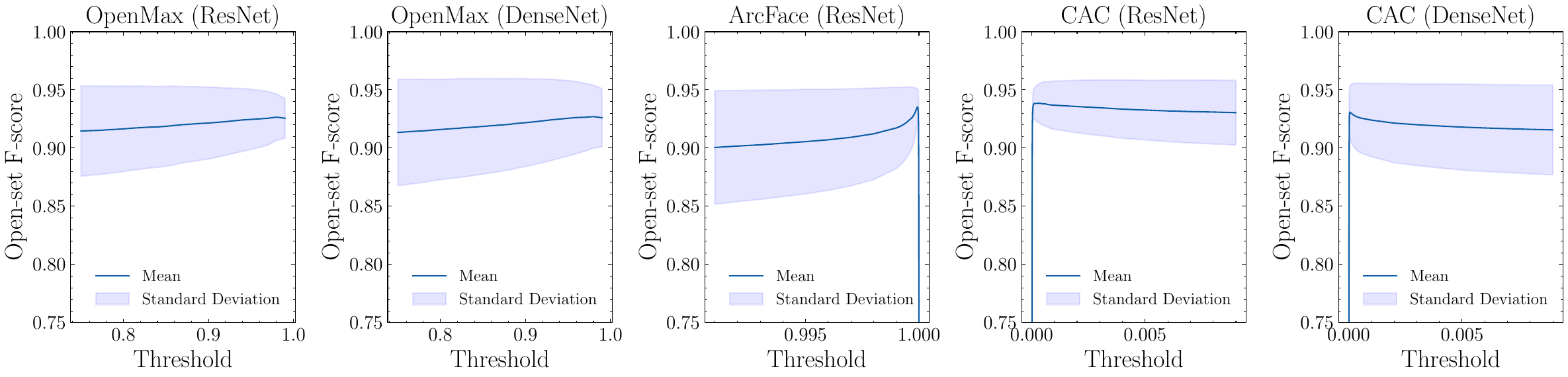}
    \caption{Average open-set F-scores with standard deviations from 5 experiments on the SYKE-plankton\_IFCB\_2022 validation set. The plots show the average performance of the baseline models across various thresholds.}
    \label{fig:thresholds_phyto_one_th}
\end{figure}

\begin{figure}[tb]
    \centering
    \includegraphics[width=0.9\linewidth]{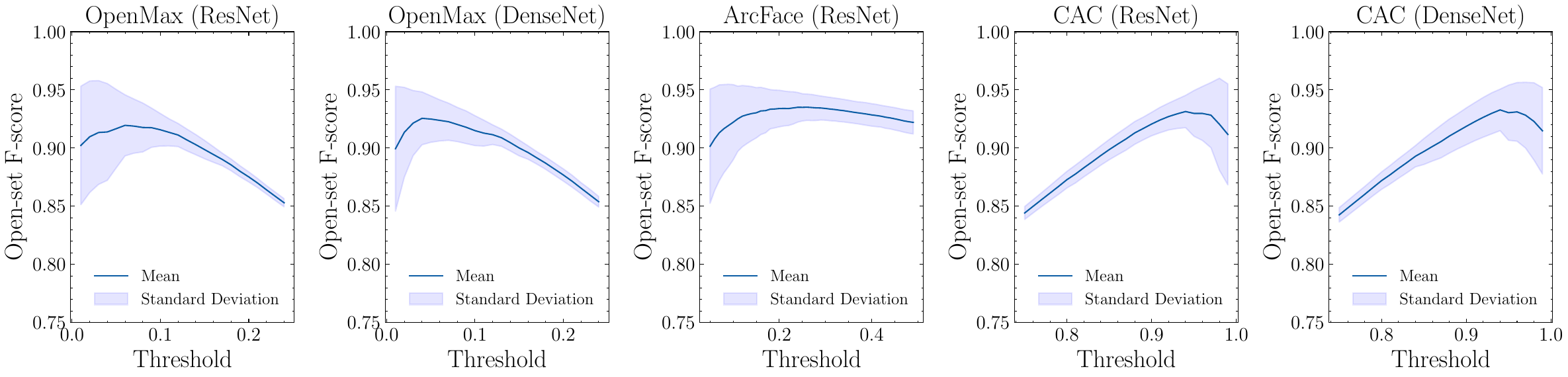}
    \caption{Average open-set F-scores with standard deviations from 5 experiments on the SYKE-plankton\_IFCB\_2022 validation set. The plots show the average performance of the baseline models across various quantiles.}
    \label{fig:thresholds_phyto}
\end{figure}

Figs.~\ref{fig:thresholds_phyto_one_th} and \ref{fig:thresholds_phyto} show how the threshold and quantile affect the open-set F-score of the validation set for the phytoplankton dataset. Similar to the thresholds for the zooplankton dataset, the results with one threshold show that OpenMax does not have a clear optimal threshold, unlike ArcFace and CAC. However, with this dataset, the peak is not as distinct for CAC utilizing ResNet. The class-specific thresholds show clear peaks for other methods except for ArcFace.  

\section{Conclusion}

In this paper, we studied open-set plankton recognition. When applied operationally, plankton imaging instruments capture various non-plankton particles and plankton species not present in the training set. This creates a challenging and environmentally relevant fine-grained OSR problem, providing a good context to develop and test OSR methods. We compared three different OSR methods on both phytoplankton and zooplankton datasets, and further analyzed the effect of the rejection threshold. We proposed setting class-specific thresholds based on the selected quantile of the scores, leading to only one tunable parameter despite a large number of individual threshold values. The CAC method, not previously applied to plankton recognition, generally provided the best accuracy. Setting class-specific thresholds increased the accuracy for most methods and datasets, but the difference was not significant, promoting the use of fixed thresholds when example images from unknown classes are available to optimize the threshold value. Despite high open-set F-scores, the accuracy of classifying unknowns leaves room for improvement. This calls for novel methods for fine-grained open-set recognition. To promote this, we have made our data publicly available~\cite{zooplanktondata}.

\section*{Acknowledgments}
The research was carried out in the FASTVISION and FASTVISION-plus projects funded by the Academy of Finland (Decision numbers 321980, 321991, 339612, and 339355).
We wish to acknowledge CSC – IT Center for Science, Finland, for computational resources.
We also acknowledge Autozoo-project (Lehtiniemi, funded by Ministry of Environment) for support in zooplankton image analysis and FINMARI as the study utilized SYKE Marine Ecological Research Laboratory infrastructure as a part of the national FINMARI RI consortium.

% ---- Bibliography ----
\bibliographystyle{splncs04}
\bibliography{main}
\end{document}